\documentclass[sigconf]{acmart}
\usepackage{subfigure}
\newcommand{\red}[1]{\textcolor{red}{#1}}
\newcommand{\grey}[1]{\textcolor{gray}{#1}}
\AtBeginDocument{%
  \providecommand\BibTeX{{%
    \normalfont B\kern-0.5em{\scshape i\kern-0.25em b}\kern-0.8em\TeX}}}

\setcopyright{acmcopyright}
\copyrightyear{2022}
\acmYear{2022}
\setcopyright{acmcopyright}
\acmConference[MM '22] {Proceedings of the 30th ACM International Conference on Multimedia }{October 10--14, 2022}{Lisboa, Portugal.}
\acmBooktitle{Proceedings of the 30th ACM International Conference on Multimedia (MM '22), October 10--14, 2022, Lisboa, Portugal}
\acmPrice{15.00}
\acmISBN{978-1-4503-9203-7/22/10}
\acmDOI{10.1145/3503161.3547799}

\begin{document}

\title{Rotation Invariant Transformer for Recognizing Object in UAVs}





\author{Shuoyi Chen}
\email{chenshuoyi@whu.edu.cn}
\affiliation{%
  \institution{School of Computer Science, Wuhan University}
  \city{Wuhan}
  \country{China}
}

\author{Mang Ye}
\email{yemang@whu.edu.cn}
\authornotemark[0]
\authornote{Corresponding author.}
\affiliation{%
  \institution{School of Computer Science, Wuhan University\\ Hubei Luojia Laboratory}
  \city{Wuhan}
  \country{China}
  }

\author{Bo Du}
\email{dubo@whu.edu.cn}
\affiliation{%
  \institution{School of Computer Science, Wuhan University\\ Hubei Luojia Laboratory}
  \city{Wuhan}
  \country{China}
  }




\renewcommand{\shortauthors}{Shuoyi Chen, Mang Ye, \& Bo Du}

\begin{abstract}
  Recognizing a target of interest from the UAVs is much more challenging than the existing object re-identification tasks across multiple city cameras. The images taken by the UAVs usually suffer from significant size difference when generating the object bounding boxes and uncertain rotation variations. Existing methods are usually designed for city cameras, incapable of handing the rotation issue in UAV scenarios. 
  A straightforward solution is to perform the image-level rotation augmentation, but it would cause loss of useful information when inputting the powerful vision transformer as patches. This motivates us to simulate the rotation operation at the patch feature level, proposing a novel rotation invariant vision transformer (RotTrans). This strategy builds on high-level features with the help of the specificity of the vision transformer structure, which enhances the robustness against large rotation differences. In addition, we design invariance constraint to establish the relationship between the original feature and the rotated features, achieving stronger rotation invariance. Our proposed transformer tested on the latest UAV datasets greatly outperforms the current state-of-the-arts, which is 5.9\% and 4.8\% higher than the highest mAP and Rank1. Notably, our model also performs competitively for the person re-identification task on traditional city cameras. In particular, our solution wins the first place in the UAV-based person re-recognition track in the Multi-Modal Video Reasoning and Analyzing Competition held in ICCV 2021. Code is available at https://github.com/whucsy/RotTrans.
\end{abstract}





\maketitle

\section{Introduction}

\begin{figure}[h]
\centering
  \includegraphics[width=7cm]{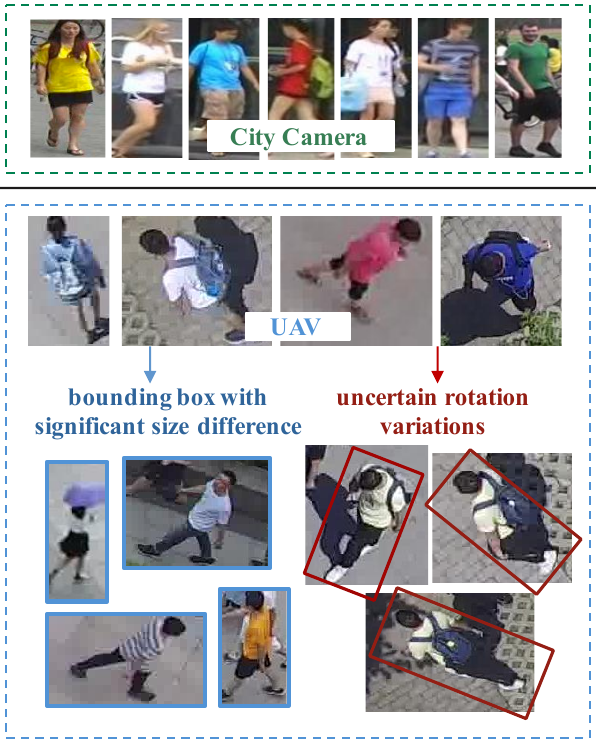}
  \caption{\small{\textbf{Images taken by city cameras \textit{vs.} Images taken by UAVs.} Recognizing person in UAV images faces two main challenges: \textit{significant size difference} and \textit{uncertain rotation variations.} }}
  \vspace{-4mm}
\label{fig:comparison}
\end{figure}

Object re-identification (ReID) is a task of retrieving specific objects such as pedestrians, vehicles across non-overlapping cameras \cite{ye2021deep,chen2017person,zheng2016person}. A large number of existing ReID researches mainly focus on city cameras. However, the commonly used city cameras have several limitations in collecting images, especially in large open areas. Specifically, the position of city cameras are fixed, which results in a limited shooting range and some blind areas \cite{zhang2020person}. With the rapid development of Unmanned Aerial Vehicles (UAVs) in the field of video surveillance, UAVs can easily cover large and hard-to-reach areas, presenting more diverse yet irreplaceable viewpoints \cite{li2021uav,kumar2020p}. The technique can be applied in various scenarios such as urban security, large-scale public place management. In this paper, we define a new task that is more challenging than normal ReID: Object recognition in UAVs, which is to recognize specific object among many aerial images captured from the moving bird-eye view.

With the emergence of ReID in UAVs, several UAV-based ReID datasets such as PRAI-1581 \cite{zhang2020person}, VRAI\cite{wang2019vehicle} and UAV-Human \cite{li2021uav} have been newly published, which promote the progress of this field. Compared to the fixed city cameras shown in Fig.~\ref{fig:comparison}, UAV moves rapidly and the altitude of it always changes, resulting in great differences in visual angle in the captured images \cite{li2021uav}. In order to recognize the identity, it is necessary to include the complete body of the target. However, it faces two main problems: 1) The shape of the generated bounding box varies greatly. It shows that the bounding box contains more background areas than that in the normal viewing angle, which makes the model easier to be affected because of the interference of some meaningless content. 2) The body of the same identity in the bounding box has varying rotation directions. This leads to a much large intra-class distance than the traditional ReID. It is challenging for the widely-used CNN models to identify objects with large rotation angles. Therefore, it is crucial to design a rotation-invariant feature learning method for UAV-ReID to solve the above problems. 

A large number of convolutional neural networks (CNN) based ReID methods such as PCB \cite{sun2018beyond}, MGN \cite{wang2018learning}, BagTricks \cite{luo2019strong}, OSNet \cite{zhou2019omni}, AGW \cite{ye2021deep} have achieved great success in the city camera scenarios, but they are sensitive to rotation in the scenario of UAVs. Some recent works \cite{he2021transreid,sharma2021person} are also competitive in applying vision transformer \cite{dosovitskiy2020image} to the field of ReID. Worse still, as mentioned above, the UAV pedestrian image unavoidable contains large portion of background, while the convolution of CNN is a typical local operation between neighboring pixels \cite{wang2018non}. As a result, the CNN-based methods usually spend too much effort on the background, which cannot explicitly model the informative target area, limiting their applicability in UAV scenario. In comparison, the vision transformer \cite{dosovitskiy2020image} has shown strong ability to model the global and long-range relationship between each input image patches parts. This property motivates us to investigate a rotation-invariant solution under the Transformer framework.
\begin{figure}[t]
\centering
  \includegraphics[width=8cm]{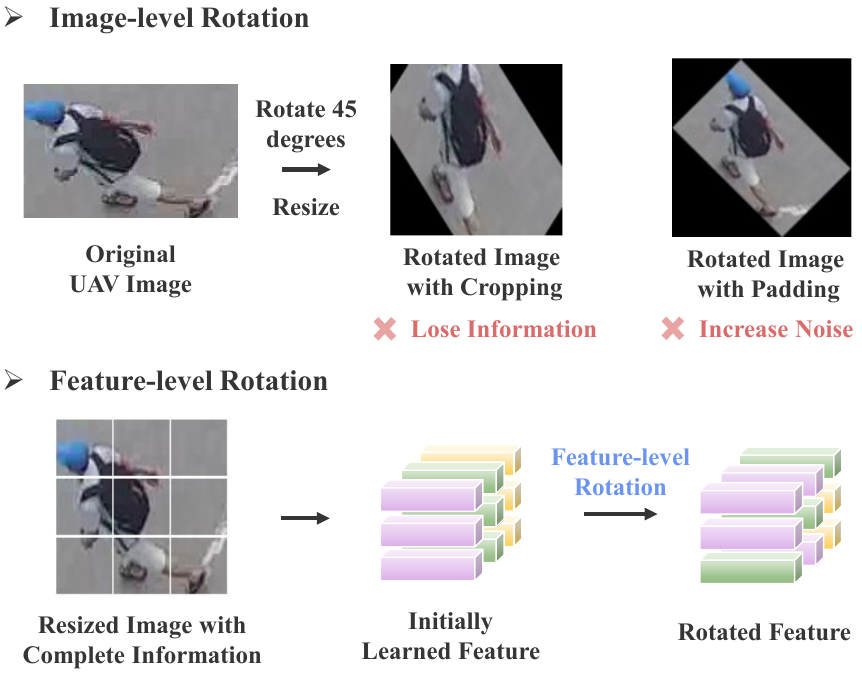}
  \caption{\small{\textbf{Image-level rotation \textit{vs.} Feature-level rotation.} \textit{Image-level rotation} would unavoidably lose discriminative information (for cropping) or introduce additional noise (for padding), while the \textit{feature-level rotation} keeps all the information.}}
  \vspace{-4mm}
\label{fig:motivation}
\end{figure}

Some existing researches based on CNN to achieve rotation invariance are applied to image classification, object detection and other visual tasks \cite{azulay2018deep, jaderberg2015spatial, cheng2016rifd}. STN \cite{jaderberg2015spatial} is proposed to improve the adaptability to image transformation by inserting learnable modules into CNN. RIFD-CNN \cite{cheng2016rifd} achieves rotation invariance by forcing the training samples before and after the rotation to share similar feature representations. For the characteristics of rotation symmetry in biological and medical images, Dieleman \textit{et al.} \cite{dieleman2016exploiting} introduce four operations as layers and inserts them into the neural network model, and make these models partially equivariant to rotations through combination. However, these methods based on convolution and two-dimensional image level operations are difficult to apply to transformers, due to the patching operations.

This paper proposes a rotation-invariant feature learning method based on vision transformer \cite{dosovitskiy2020image}, which can overcome rotation challenge of object recognition in UAVs. A straightforward solution is to perform the random rotation augmentation at the input image level by simulating rotation variations. However, this simple image-level rotation will not work under the Transformer framework. As shown in Fig.~\ref{fig:motivation}, 
there are two major drawbacks for image-level rotation operations, where we need cropping or padding strategy to match the fixed image size input. However, the cropping would lose important discriminative information and padding introduces more background noise,  which further exacerbate the challenges faced by ReID in UAVs as mentioned in Fig.~\ref{fig:comparison}.
Instead, we design a rotation strategy at the feature level. The basic idea is to simulate rotation operations on initially learned features to introduce rotation diversity. This method can alleviate the information loss caused by rotation, because as the network depth increases, the model has the ability to integrate global information of the original image \cite{dosovitskiy2020image}, after which the feature level rotation operation is performed. The experiment on the comparison of image-level rotation and feature-level rotation is in \S.~\ref{ablation}. In addition, due to the randomness of the rotation, it may result in an ambiguity issue for feature representation. We further incorporate a regularization constraint to enhance the invariance by constraining the close relationship between features of the same identity under varying rotations. Our main contributions are summarized as follows: 
\begin{itemize}
\item We analyze the difficulty of object recognition task in UAV scenarios and propose a rotation invariant vision transformer (RotTrans) which designs a feature-level rotation strategy to enhance the generalization against rotation variations. 
\item We integrate the rotation invariance constraint into the feature learning process, enhancing the robustness against spatial changes. This also reduces incorrect classification caused by rotation transformation.
\item We evaluate our method on both UAV and city-camera based datasets, achieving much better performance than the state-of-the-arts. Notably, on the challenging PRAI-1581 dataset, the Rank-1/mAP is improved from 63.3\%/55.1\% to 70.8\%/63.7\%.
\end{itemize}

\section{Related Work}
\textbf{Object Re-Identification in City Cameras.} With the in-depth development of person and vehicle re-identification research, many important breakthroughs have been made in this field \cite{chen2020salience, fan2020learning, jin2020style, chen2021neural, wang2020high, zhang2020relation, pu2021lifelong, meng2020parsing, chen2022structure, ye2021collaborative, li2021weperson, ye2021channel}. Many existing methods focus on the convolutional neural network model to improve feature representation learning \cite{ye2021deep, xiao2016learning, zheng2017person}. Luo \textit{et al.} \cite{luo2019strong} propose a strong baseline based on CNN, which can achieve good performance only by using a global feature and adding some tricks. In addition, there are some ReID methods that use part-level features. PCB \cite{sun2018beyond} provides a baseline for learning local features. To provide fine-grained local feature representation, a parsing-based view-aware embedding network \cite{meng2020parsing} is proposed. The attention mechanism has also been extensively studied in the ReID \cite{li2018harmonious, si2018dual, shen2018end, yang2019attention, wang2018mancs, chen2019self}. Recently, more and more work have applied Transformer to the field of computer vision, such as DETR \cite{zhu2020deformable}, ViT \cite{dosovitskiy2020image}, DeiT \cite{touvron2021training}, T2T-ViT \cite{yuan2021tokens}, PVT \cite{wang2021pyramid}. TransReID \cite{he2021transreid} is the first work to use pure Transformer for ReID research. It achieves good performance on both person and vehicle ReID tasks by improving the vision transformer. The following research \cite{wang2021pose} is based on TransReID, by using the pose information and introducing the transformer decoder, to solve the occlusion problem in the ReID task. Compared with CNN, Transformer has more powerful modeling capabilities in modeling the relationship between all patches and capture long-distance dependencies \cite{liu2021swin, chi2020relationnet++}.

\begin{figure*}
\centering
\includegraphics[width=16.7cm]{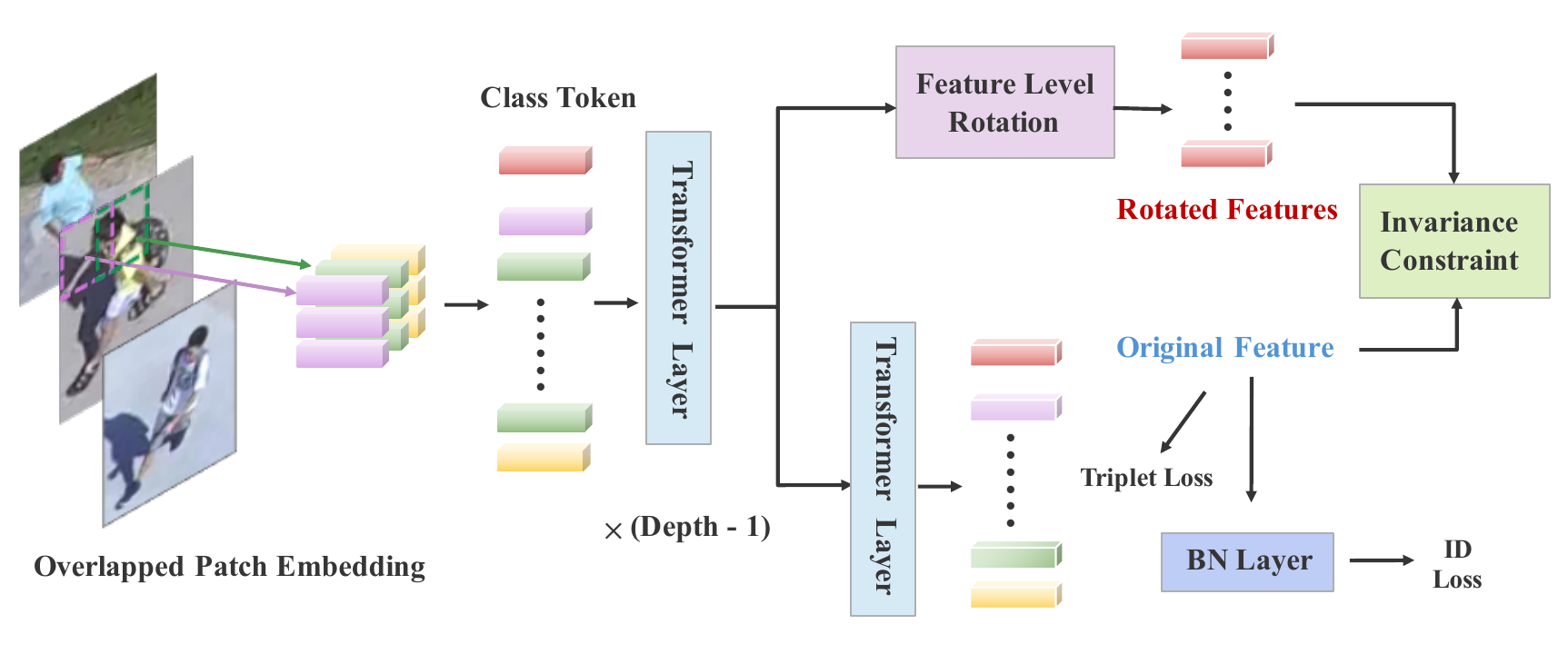} 
\vspace{-3mm}
\caption{\small{\textbf{The architecture of our proposed Rotated Vision Transformer.} The feature-level rotation strategy enhances the generalization ability of the rotation transformation by generating and learning multiple rotated features. In the invariance constraint strategy, the rotated features are used to establish constraints with the original features learned by the vision transformer as the backbone.}} 
\label{fig:architecture}
\end{figure*}

\textbf{Object Re-Identification in UAVs.}
Object Re-Identification in UAVs has attracted attention in recent years. DRone HIT \cite{grigorev2019deep} is a data set of 101 identities collected by UAVs, which is used for an earlier attempt of person ReID under UAV conditions. Zhang \textit{et al.} \cite{zhang2020person} propose a large Person ReID dataset PRAI-1581 in real UAV surveillance scenarios and utilize the subspace pooling layer to get compact feature representations. VRAI \cite{wang2019vehicle} is a vehicle ReID dataset captured by UAV. A multi-task learning framework that utilizes manually labeled auxiliary information such as vehicle type and color is presented. But the methods they proposed can also be applied to common scenarios, not for the unique problems of UAV scenarios. P-DESTRE \cite{kumar2020p} is a video-based UAV dataset that can perform various tasks such as pedestrian detection, tracking, and re-identification. Li \textit{et al.} \cite{li2021uav} present a large-scale UAV-Human dataset for human behavior understanding with UAVs. Both of them \cite{kumar2020p, li2021uav} just provide datasets without designing methods for person ReID task. In comparison, this paper studies the important rotation and varying image sizes problems in UAVs, which are ignored in existing methods.

\textbf{Rotation Invariance.}
Objects under the view of UAVs usually have multiple rotation angles. The existing object re-recognition methods do not take the rotation invariance into account, because this problem does not exist in the city camera scene. However, there are some research on rotation invariance in tasks of image classification and object detection \cite{jaderberg2015spatial, cheng2016rifd, zhang2021vit}. The CNN structure itself does not have strict invariance, even a small pixel change may bring a big difference in detection results \cite{azulay2018deep}. Jaderberg \textit{et al.} \cite{jaderberg2015spatial} propose to integrate the spatial transformer layer into the CNN. It allows the network to automatically learn how to perform spatial changes in the feature map, reducing the overall cost of training. But this convolution-based layer cannot be applied to the vision transformer. RIFD-CNN \cite{cheng2016rifd} uses data augmentation to obtain multiple rotated images to expand the training set, and optimize a new objective function to share similar features before and after the rotation of images. But image rotation brings about pixel changes and results in some loss of information. Besides, using rotation detection \cite{ding2019learning, yang2021r3det,  han2021redet, yang2021rethinking, yang2021learning, yang2022arbitrary} to generate the object bounding boxes can be an effective strategy to reduce the background noise problem. However, the re-detected rotated bounding box might be non-vertical and cannot match the vertical rectangular input of the network. It is difficult to obtain a vertical rectangular bounding box without destroying the pedestrian body structure in the correct orientation.

\section{Proposed Method}
The goal of our designed rotated transformer (RotTrans) is to learn rotation-invariant feature representations based on images captured by UAVs. The vision transformer has powerful modeling ability and generalization ability \cite{naseer2021intriguing}, which makes an excellent performance on common object recognition tasks. In order to make the model able to cope with the challenge of rotation and utilize the effective advantages of vision transformer, our method is built on the vision transformer with two main components: 1) \textit{Feature Level Rotation}: due to the limitations of direct rotation at the image level, we designed an operation that simulates the rotation of patch features at the feature level to generate diverse rotated features. 2) \textit{Invariance Constraint}: establish strong constraints between multiple rotated features and original features, jointly optimizing with the identity-invariant target. Notably, the two strategies we propose above can be embedded into any existing transformer architecture without affecting the original network structure. This idea can be widely used in different types of tasks as a solution to achieve rotation invariance.
The overall network architecture is shown in Fig.~\ref{fig:architecture}.

\subsection{Vision Transformer Backbone}
Our architecture follows the transformer baseline for the ReID task proposed by He \textit{et al.} \cite{he2021transreid}, which improves the backbone of the vision transformer. The detail will be introduced in this part. For an image $\mathbf{x} \in \mathbb{R}^{H \times W \times C}$, where $H \times W$ represents the image resolution and $C$ represents the three channels of the RGB image, the vision transformer divides it into $N$ patches through the patch embedding operation, and the size of each patch is denoted by $P \times P$ during the division. Then, the patches are projected to the space of dimension $D$ which is $P^2 \times C$ through linear transformation. The final input is composed of N one-dimensional vectors, denoted as $\mathbf{x} \in \mathbb{R}^{N \times D}$. Due to the limitation of this direct division method that is difficult to learn the internal information of each patch, overlapping patch embedding is adopted. The specific implementation is using a convolutional layer with kernel size of $P$ and the stride size of $S$, where $S<P$. In addition, the learnable embedding (class token) is added to the patch embedding as a vector for classification and used as the input of the Transformer together with the patch embedding. Position embedding $E_{pos} \in \mathbb{R}^{(N+1) \times D}$ is also added to patch embedding to preserve the spatial position information, which is also learnable. The class token (represented by $c$) obtained after backbone learning is used as a global feature representation, and the triplet loss \cite{arxiv17triplet} and cross entropy loss widely used in Re-ID tasks are adopted to optimize the network.

\subsection{Feature-level Rotation}
Although the global feature already contains rich information and has strong ability to identify the target, the transformer is entirely based on the attention mechanism, which has no rotation invariance and is sensitive to scenarios for recognizing rotating objects. Enhancing the generalization ability of the transformer for rotation changes in the UAV scenario is a critical challenge. The data augmentation of simple and direct random rotation at the image level is not applicable in the vision transformer. Due to the defects of both strategies of direct rotation operation at the image level, cropping will cause the loss of effective information and padding will increase the background noise. The rotated image is further divided into patches as the input of the transformer, which increases the negative impact. To ensure that the model can learn better feature representations and achieve rotation invariance while maintaining the complete information of the original image, we propose a novel feature-level rotation. By simulating multiple rotation transformations on patches at the feature level, the prior knowledge of the rotation transformation is introduced in the training process. This strategy guarantees the generalization ability while preventing the valid information from being destroyed. Furthermore, the feature-level rotation design can be embedded into any existing transformer architecture without affecting original updating and network structure. The specific implementation steps are as follows:

\textbf{Reshape.} The global feature learned by backbone is denoted as $f \in \mathbb{R}^{(N+1) \times D}$, where $N+1$ represents a patch sequence with length $N$ (denoted as $f_p$) and a class token (denoted as $c_o$). In order to simulate the rotation operation in two-dimensional space, we reshape $f_p \in \mathbb{R}^{N \times D}$ to $f_{res} \in \mathbb{R}^{X \times Y \times D}$ where $(X,Y)$ represents the spatial size of patches generated by overlapping patch embedding with a stride size of $S$. The calculation formulas of $X$ and $Y$ are:
\vspace{-1mm}
\begin{equation}
X = \lfloor{\frac{H - P}{S}}\rfloor + 1, Y = \lfloor{\frac{W - P}{S}}\rfloor + 1.
\vspace{-1mm}
\end{equation}

\begin{figure*}
\centering
\includegraphics[width=16cm]{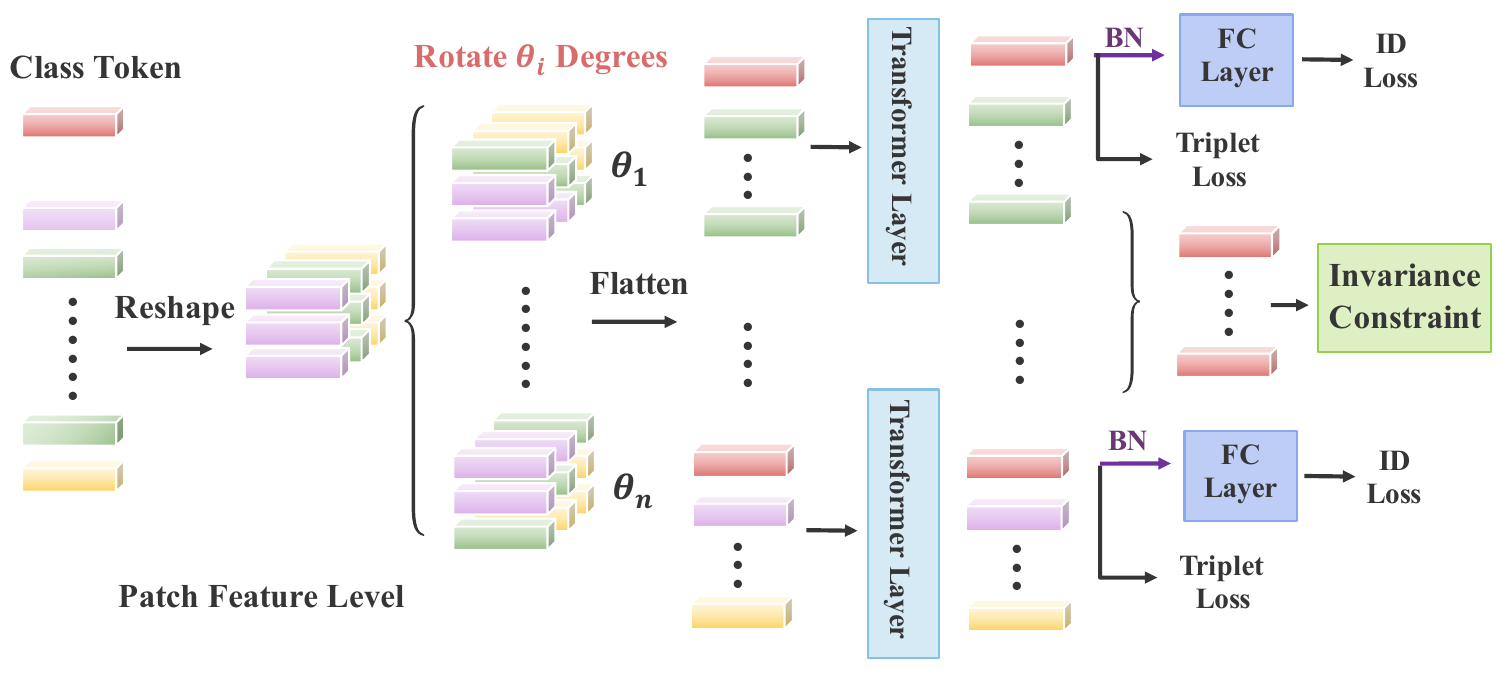} 
\caption{\small{\textbf{Details of the feature-level rotation strategy.} The simulated rotation operation is performed on initially learned patch features. The multiple rotation feature representations updated by the transformer layer enhance the generalization ability of the model.}} 
\label{fig:rotation}
\end{figure*}

\textbf{Rotated Features Generation.} We treat each patch as a pixel, $f_{res}$ can be visually regarded as a two-dimensional matrix. In this way, we can apply the operation similar to rotating matrix to the patch feature level. Due to the continuous movement of the UAV, the angle of the captured matrix changes unpredictably. We simulate this angle diversity by randomly generating a set of degrees $A= \{\theta_i|i=1,2,\cdots,n\}$. The coordinates of each patch vector in the reshaped two-dimensional space are represented by $(x, y)$, and the operation of rotating a degree of $\theta_i$ can be formulated as:
\vspace{-1mm}
\begin{equation}
\begin{split}
x^k_r = x\cos{\theta_i} - y\sin{\theta_i},\\
y^k_r = x\sin{\theta_i} + y\cos{\theta_i}.
\end{split}
\vspace{-1mm}
\end{equation}

Unlike pixel-based image rotation, the feature-level rotation is performed on larger patches, so that rotating a degree that looks small in number actually simulates a relatively large rotation. Therefore, we define a variable $\alpha$ to constrain the size of the generated degrees, which is $\theta_i \in [-\alpha,\alpha]$. As shown in Fig.~\ref{fig:rotation}, a set of multi-angle rotated features $F_{r} = \{f_{r1},f_{r2},\cdots,f_{rn}\}$ are obtained by performing the above-mentioned rotation operation. At this stage, the feature representations with various angles in the UAV scene are integrated into the model in advance to simulate rotation variation, and the class token learns the complete information of the original image.

\textbf{Model Training.} Due to the randomness of rotation, each rotated feature containing different information can be regarded as a new feature representation. In order to learn a variety of features, the two-dimensional $f_{r}\in \mathbb{R}^{X \times Y \times D}$ should be flattened to $f_{r}\in \mathbb{R}^{ N \times D}$, so that the transformer can receive the one-dimensional patch sequence. Each rotated feature has the same number of patches of size $N$ as the original feature, and there is a problem that it is difficult to include the information of all patches when used for classification. The class token learned by multiple transformer layers integrates the global feature representation. We obtain $f_{r}\in \mathbb{R}^{(N+1) \times D}$ by adding $n$ copies of the original feature's class token $c_o$ to each rotated feature. The purpose of this operation is that each rotated feature can be classified using a learnable class token as a representative of $n$ rotated patches. Then, we set up separate transformer layers for each of them to ensure that the diversity can be learned. The update of the class tokens representing the rotated features $c_r$ during training is based on the original class token $c_o$ that already contains rich feature information, which effectively avoids the loss caused by feature rotation. We set up $n$ classifiers for the class tokens of the rotated feature updated by the transformer layer. The most commonly used cross entropy loss ($\mathcal{L}_{CE}$) is adopted after the batch normlization layer \cite{luo2019strong}. Furthermore, triplet loss \cite{arxiv17triplet} ($\mathcal{L}_{T}$) is used in each class token for fine-grained recognition of objects. This final loss for rotated features is formulated as:
\vspace{-1mm}
\begin{equation}
{\mathcal{L}_{rot}}(c_{ri}) = \frac{1}{n}\sum_{i=1}^n (\mathcal{L}_{T}(c_{ri})+\mathcal{L}_{CE}(c_{ri})).
\vspace{-1mm}
\end{equation}
Each class token representing rotated feature plays equal contribution in the overall model update.

\subsection{Invariant Constraint}
The feature level rotation module in the previous section improves the model generalization ability for angle transformation from the perspective of diversity. In addition, the rotated features and the original feature both represent the same pedestrian. We artificially incorporate additional invariance constraints for rotated features and the original feature to the loss function to establish their correlation between them \cite{ye2020augmentation}. In this way, the distance within the class is shortened, which is more conducive to correct classification.

There is a many-to-one relationship between a group of class tokens $C_{r} = \{c_{r1},c_{r2},\cdots,c_{rn}\}$ of the rotated features and a class token $c^{\prime}_o$ of the original feature. A large computational cost will be generated if the constraint is established for each pair of original feature and each rotated feature. To avoid redundant computation, the average of rotated features is used to establish the invariance constraint, which is denoted by
\begin{equation}
\bar{c_r} = \frac{1}{n}(c_{r1}+c_{r2}+\cdots+c_{rn}).
\vspace{-1mm}
\end{equation}
Our goal is to constrain the difference between the average rotated feature and the original feature. It is necessary to ensure that the class discrimination of the rotated feature representation will not be weakened. Mean Square Error (MSE) is the most commonly used loss function, it represents the sum of squares of the difference between the predicted value and the target value. We choose the smooth L1 Loss to calculate the difference, which can effectively prevent the problem of gradient explosion. The rotation invariance constraint of this part is then formulated as:

\vspace{-1mm}
\begin{equation}\label{eq:indicator}
\mathcal{L}(c^{\prime}_o,\bar{c_r})=\left\{
\begin{aligned}
0.5 {(c^{\prime}_o - \bar{c_r})}^2, & \quad \mid c^{\prime}_o - \bar{c_r}\mid < 1 \\
\mid c^{\prime}_o - \bar{c_r}\mid-0.5, & \quad otherwise.
\end{aligned}
\right.
\vspace{-1mm}
\end{equation}

In the training phase, the overall loss function consists of three parts. When the rotated features are updated, the original feature is also input to a transformer layer to get a further updated class token as the global representation. We denote the original feature representation learned through multiple transformer layers as $c^{\prime}_o$. The triplet loss is also adopted and the cross entropy loss is adopted after the batch normlization layer:
\vspace{-1mm}
\begin{equation}
{\mathcal{L}_{ori}}(c^{\prime}_o) = \mathcal{L}_{T}(c^{\prime}_o)+\mathcal{L}_{CE}(c^{\prime}_o).
\vspace{-1mm}
\end{equation}
Besides, the average rotated feature is an auxiliary feature representation that adapts to the diversity of angles. The invariance constraint controls the difference between the original feature and the rotated features. In summary, the overall learning objective is 
\vspace{-1mm}
\begin{equation}
\begin{split}
\mathcal{L} = \lambda\mathcal{L}_{ori}(c^{\prime}_o) + (1 - \lambda)\mathcal{L}_{rot}(c_{ri}) + \mathcal{L}(c^{\prime}_o,\bar{c_r})
\end{split}
\vspace{-1mm}
\end{equation}
where $\lambda$ and $1-\lambda$ respectively represent the weights of original features and rotated features.

\begin{table*}[t]
\centering
 \caption{\label{tab:test-uav}\small{Evaluation of the current state-of-the-arts on two person datasets (PRAI-1581 and UAV-Human) and one vehicle dataset captured by UAVs. The performance of our proposed Rotated Vision Transformer is significantly improved compared to other methods, especially in PRAI-1581 which is more challenge. Rank1 and Rank5 accuracy (\%), mAP (\%) and mINP (\%) are reported. }}
\begin{tabular}{p{2.5cm}p{2cm}p{0.8cm}p{0.8cm}p{0.8cm}p{0.8cm}p{0.8cm}p{0.8cm}p{0.8cm}p{0.8cm}p{0.8cm}p{0.8cm}p{0.8cm}}
\toprule
\multicolumn{2}{c}{Dataset} &\multicolumn{4}{c}{PRAI-1581}  & \multicolumn{4}{c}{UAV-Human} & \multicolumn{3}{c}{VRAI}\\
\midrule
Method &Venue & Rank1  & Rank5 & mAP  & mINP  & Rank1  & Rank5  & mAP   & mINP  & Rank1  & Rank5  & mAP   \\
\midrule
\multicolumn{7}{l}{\textit{(a) Evaluation of CNN based methods}} \\
\midrule
PCB \cite{sun2018beyond} & ECCV-18 & 47.5 &   - & 37.2 & - &  62.2 &  83.9 &  61.1 & - & - & -& - \\
MGN \cite{wang2018learning} & ACM-18 & 49.6 &   - & 40.9 & - &  - & - & - & - & 67.8 & 82.8 & 69.5 \\
DG-Net \cite{zheng2019joint} & CVPR-19 & - &   - & - & - &  65.8 &  85.7 &  62.0 & - & - & -& - \\
OSNET \cite{zhou2019omni} & ICCV-19 & 54.4 &   - & 42.1 & - &  - & - & - & -& - & -& - \\
BagTricks \cite{luo2019strong} & TMM-19 & 50.1 &   67.3 & 42.5 & 19.9 &  61.6 & 84.8 & 62.1 & 52.6 & - & -& -\\
SP\cite{zhang2020person} & TMM-20 &  54.8 &  -& 43.1  & - &  - & - & - & - & - & -& - \\
AGW \cite{ye2021deep} & TPAMI-21 & 53.0 &  69.3 & 45.6 & 22.9 &  66.2 & 86.2 &66.6 & 57.8 & 67.6 & 79.9& 68.1 \\
Multi-task\cite{wang2019vehicle} & ICCV-19 & -& - & - & - & -& - & -& - & 80.3 & 88.5 & 78.6 \\
\midrule
\multicolumn{7}{l}{\textit{(b) Evaluation of Transformer based methods}}\\
\midrule
Baseline (VIT) \cite{he2021transreid}& ICCV-21 & 63.3 & 78.7 &  55.1 & 30.3 &  72.0 & 89.4 & 71.7 & 64.2& - & -& - \\
TransReID \cite{he2021transreid}  & ICCV-21 & 66.0 &  79.6 & 57.8 & 33.1 & 73.6 &  89.0 & 73.0 & 66.3 & 81.0 & 89.8& 82.37 \\
\midrule
RotTrans (Ours)  & - &  \bf{70.8} &	\bf{84.1}	& \bf{63.7} & \bf{39.6} & \bf{75.6} &	\bf{90.1} & \bf{74.4}	& \bf{67.8}& \bf{83.5} & \bf{90.9}& \bf{84.8} \\
\bottomrule
 \end{tabular}
   \vspace{-2mm}
\end{table*}

\section{Experiments}
\subsection{Object ReID in UAVs}
\textbf{Datasets.} We evaluate our method on two recently published person ReID datasets and one vehicle datasets (PRAI-1581 \cite{zhang2020person}, UAV-Human \cite{li2021uav} and VRAI \cite{wang2019vehicle}) collected by UAVs and two widely-used datasets (Market-1501 \cite{zheng2015scalable} and MSMT17\cite{wei2018person}) captured by general city cameras. Due to the continuous changes in altitude, angle, and environment during the flight of UAVs, these images are more complicated than the images captured by ordinary city cameras. Regarding the privacy of person images in the dataset, the data collection agreements are signed during the collection process. Cumulative Matching Characteristic (CMC) and the mean Average Precision (mAP), the two most commonly used metrics in ReID tasks are used for our evaluation. In addition, we also adopt a new evaluation metric mean Inverse Negative Penalty (mINP) \cite{ye2021deep}, as a supplement to CMC/mAP, which indicates the cost of finding the hardest matching sample.

PRAI-1581 dataset \cite{zhang2020person} is proposed for person ReID task. It is collected by two UAVs flying at the altitude of 20 to 60 meters, including 39,461 images of 1581 person identities. It is a very typical UAV scene from a bird's-eye view. The traditional ReID task is a cross-camera retrieval task, and the annotation information of the camera is required for evaluation. The author uses the same method as the traditional city camera scene when labeling the dataset, and each UAV represents a camera number. However, for a fixed city camera, the background and shooting angle of view are static. The images taken by the same camera of the UAV are always in a dynamically changing environment. In fact, Most images can be considered as cross-camera scenes.

UAV-Human dataset \cite{li2021uav} is mainly used for human behavior understanding in UAVs, and can be applied to various tasks such as person ReID, action recognition, and attitude estimation. The dataset contains 41,290 pedestrian images and 1,144 identities. However, the flying height of the UAV is relatively low, between 2 and 8 meters, which is easier to be correctly identified. Its main difficulty lies in the presence of some images in low-light conditions. Notably, it also involves dynamic environmental changes.

VRAI dataset \cite{wang2019vehicle} is constructed for vehicle ReID task which consists of 137,613 images of 13,022 vehicle instances. Vehicle pictures are collected by UAVs flying at altitudes of 15 to 80 meters in different places. At the same time, rich annotations are provided, including colors, vehicle types, attributes, discriminative parts of the images. This dataset is challenging because it contains diverse viewpoints, larger pose variation and wider range of resolution. 

\textbf{Implementation Details.} We divide the PRAI-1581 data set according to the division method given by the author and the UAV-Human and the VRAI are already divided. For the input image, the generated bounding box of the object taken by the UAV has various sizes and shapes, so we uniformly resize the image to 256*256. In addition, padding with 10 pixels, random cropping and random erasing with probability 0.5 are adopted in the training data. The vision transformer \cite{dosovitskiy2020image} pre-trained on imageNet-1K is used as the backbone for person feature extraction. The patch size is set to 16*16 and the stride size is set to 12*12 at the stage of overlapping patch embedding. In the feature level rotation, the number of rotated features $N$ is 4 and the range of the random rotation angle is between -15 degrees and 15 degrees. Since it is based on patch rotation, the angle should not be set too large. For the original features and rotated features extracted by the backbone, triplet loss without margin is used and the cross entropy loss is used after the features pass through the batch normalization layer. The weigth of original feature $\lambda$ is 0.5 and the weight of rotated features $1-\lambda$ is 0.5. Smooth l1 loss is applied between the average rotated feature and the original feature. During training, the Stochastic Gradient Descent (SGD) optimizer is used. The initial learning rate is 0.008, and the cosine learning rate decays is adopted. The number of training epochs is 200. The batch size is set to 64, including 16 identities, each with 4 images. In the test phase, only the original features are used to calculate the distance matrix. The experiment is implemented based on both Pytorch and Huawei MindSpore \cite{mindspore}.

\subsection{Comparison with State-of-the-Art Methods}
Table \ref{tab:test-uav} shows the comparison between our method and the existing state-of-the-arts on the two UAV datasets. The performance of our proposed Rotated Vision Transformer is better than all CNN and Transformer based ReID methods.

\begin{table}[t]
\centering
\caption{\label{tab:effet}Ablation study of Feature Level Rotation and Invariant Constraint we proposed on the PRAI-1581. Rank1, Rank5 and Rank10 accuracy (\%), mAP (\%) are reported. }
\begin{tabular}{p{3cm}p{0.9cm}p{0.9cm}p{0.9cm}p{0.9cm}}
\toprule
& \multicolumn{4}{c}{PRAI-1581}  \\
 Strategy &mAP & Rank1 & Rank5 & Rank10  \\
\midrule
Baseline (VIT) \cite{he2021transreid} & 55.1 & 63.3   & 78.7 & 85.1 \\
\midrule
\multicolumn{5}{l}{\textit{(a) Rotation operation at the image level}} \\
\midrule
+ Random Rotation & 50.2 & 58.3   & 74.8 & 81.6 \\
 & \small{\grey{$\downarrow$4.9}}&  \small{\grey{$\downarrow$5.0}} & \small{\grey{$\downarrow$3.9}} & \small{\grey{$\downarrow$3.5}} \\\hline
 \multicolumn{5}{l}{\textit{(b) Rotation operation at the feature level}} \\
\midrule
+ Feature Level Rotation &\bf{61.4} & \bf{68.9} & \bf{82.2} & \bf{87.5}\\
 & \small{\red{$\uparrow$6.3}}&  \small{\red{$\uparrow$5.6}} & \small{\red{$\uparrow$3.5}} & \small{\red{$\uparrow$2.4}} \\
+ Invariant Constraint & \bf{63.7} & \bf{70.8}  & \bf{84.1} &\bf{89.0}\\
 & \small{\red{$\uparrow$2.3}}&  \small{\red{$\uparrow$1.9}} & \small{\red{$\uparrow$1.9}}&\small{\red{$\uparrow$1.5}}\\
 \bottomrule
 \end{tabular}
  \vspace{-2mm}
\end{table}

\textbf{Result on PRAI-1581.} The results of the transformer based methods are experimentally obtained under the same settings as our method. Among the methods based on CNN, AGW\cite{ye2021deep} achieved the best mAP of 45.61\% in the backbone of ResNet50. However, the performance of the transformer methods are obviously much better, even using the simplest baseline of the vision transformer, the mAP obtained exceeds the best method of CNN by +7.45\%. It proves that the transformer has more powerful global modeling capabilities in the extreme perspective scenes of UAVs. The dataset PRAI-1581 contains a large number of pedestrian images with varying angles. Compared with the transformer methods which do not have space invariance, our method has a good adaptability against the random angle rotation. The performance of our method is significantly better than all methods in the table, achieving the highest 70.8\% Rank1 accuracy rate and 63.7\% mAP, which exceeds the best results of other methods by 5.9\%.

\textbf{Result on UAV-Human.} Transformer methods still outperforms CNN methods on this data set overall, but the improvement is not as obvious as in PRAI-1581. Mainly due to the low flight altitude when UAV-Human collects data, the angle of pedestrians in space does not change much, and the influence of rotation is small. But our method still achieves the best performance, with mAP reaching 74.4\% and rank1 reaching 75.6\%. In particular, in the first ICCV workshop competition on Multi-Modal Video Reasoning and Analyzing this year, we win the first place in the Person Re-Identification based on UAV-Human track \cite{peng2021multi}. Among the 68 participating teams from world-renowned research institutions, our team achieved 79.1\% mAP, which is 4.6\% higher than the second-ranked team. However, tricks such as re-rank and muti-shot are used in the competition, which cannot be directly compared with the results in the table. 

\textbf{Result on VRAI.} The VRAI dataset does not publish complete test data, only by submitting the specified format to the challenge they host to test the performance of the proposed method. Due to the difficulty of the vehicle ReID task is different from that of pedestrians, some methods designed for pedestrian ReID do not work well for vehicles. In the CNN method, the Multi-task with attribute annotation achieves better performance. However, the vehicle image in the UAV scene also has a variety of rotation angles. Our proposed RotTrans achieves 83.5\% Rank1 accuracy rate and 84.8\% mAP without using any auxiliary information, surpassing all other methods.

\begin{figure}[t]
\centering{
 \includegraphics[width=0.49\linewidth]{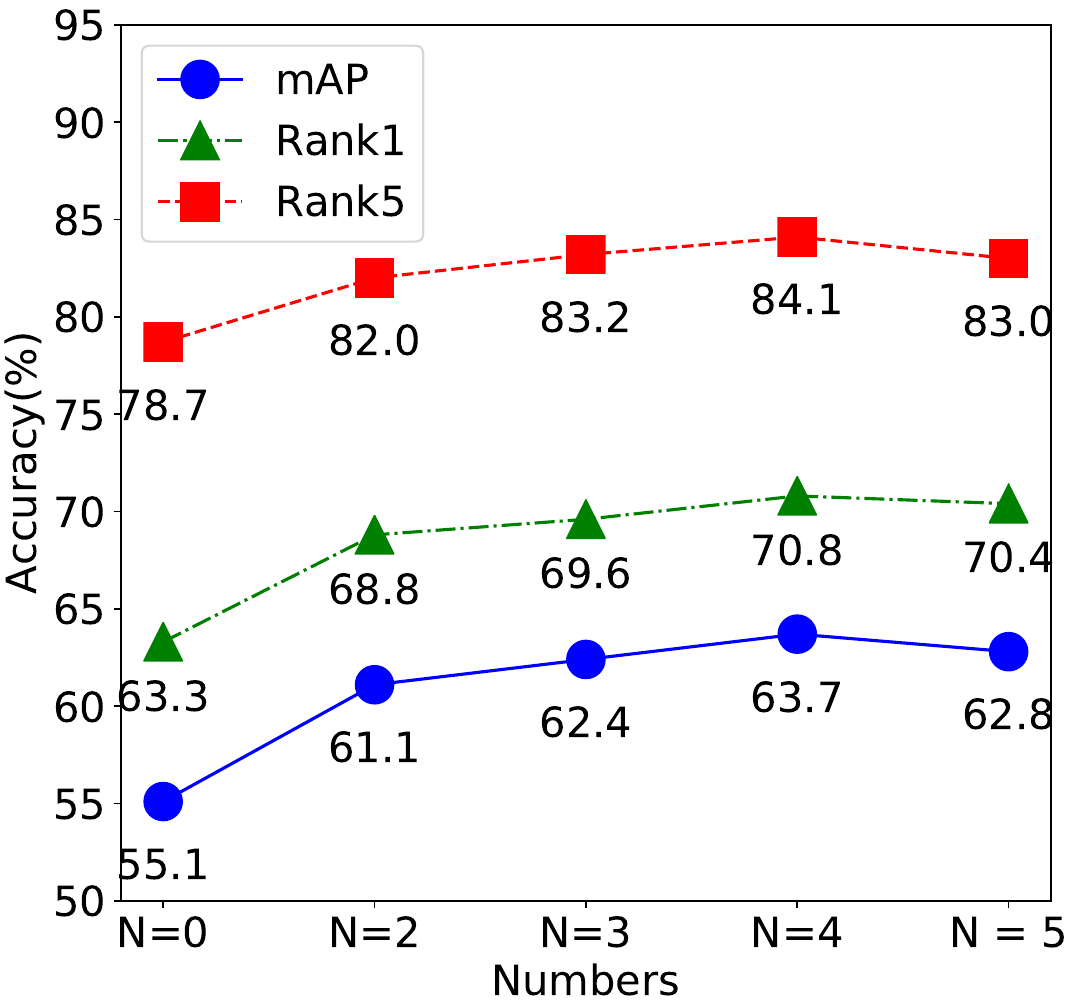} 
 \includegraphics[width=0.49\linewidth]{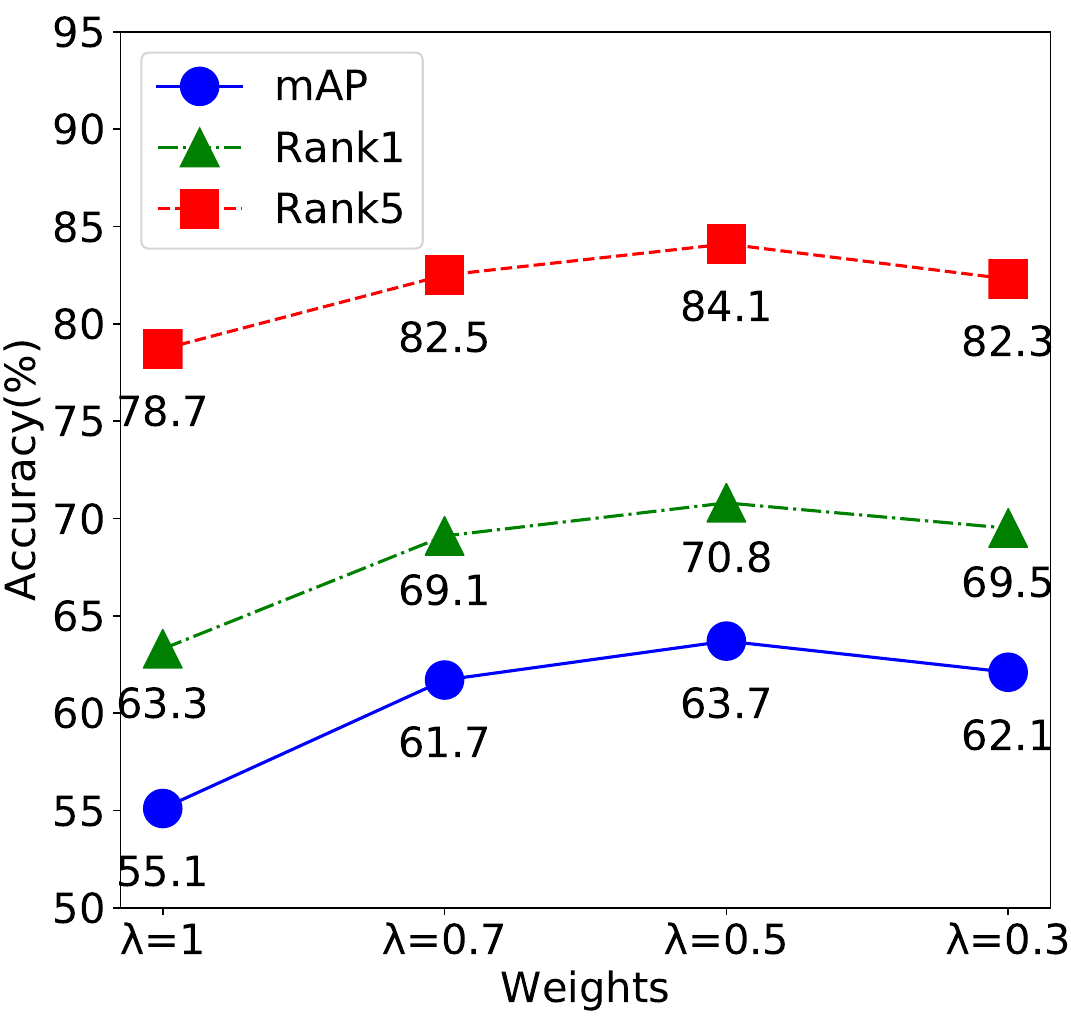}
}
\caption{Experiments with different numbers of rotated features and weights of the original feature in Feature Level Rotation. Evaluation on dataset PRAI-1581.
\label{fig:parameter}}
  \vspace{-5mm}
\end{figure}

\subsection{Ablation Study on UAV Datasets}
\label{ablation}

\textbf{Why Feature-level Rotation?} Direct rotation data augmentation is not applicable in UAV scenarios. In order to verify the negative effects of simple image-level rotation augmentation, we conducted specific experiments. As shown in Table \ref{tab:effet}, we introduce the data augmentation operation of - 30 to 30 degrees random rotation in the training set based on the vision transformer baseline, which causes mAP/Rank1 to drop by 4.9/5.0. Rotating the low-resolution image of the UAV is easy to lose the discriminative information, which makes the information learned by the model ambiguous. However, Our proposed feature-level rotation strategy effectively avoids the loss of discriminative information by integrating global information through the learning of class tokens through multi-layer transformer layers. On this basis, the rotation diversity is increased to significantly improve the model performance. Our experiments to verify the effectiveness of the feature-level rotation strategy show that mAP and rank1 accuracy have increased by 6.3\% and 5.6\% respectively compared to the baseline.

\begin{figure*}[t]
\centering{
 \subfigure[ Samples without Rotating Operation ]{
 \includegraphics[width=0.24\linewidth]{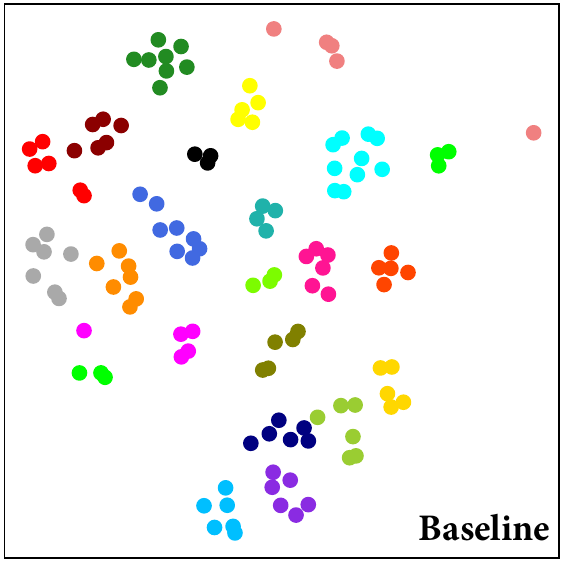} \includegraphics[width=0.24\linewidth]{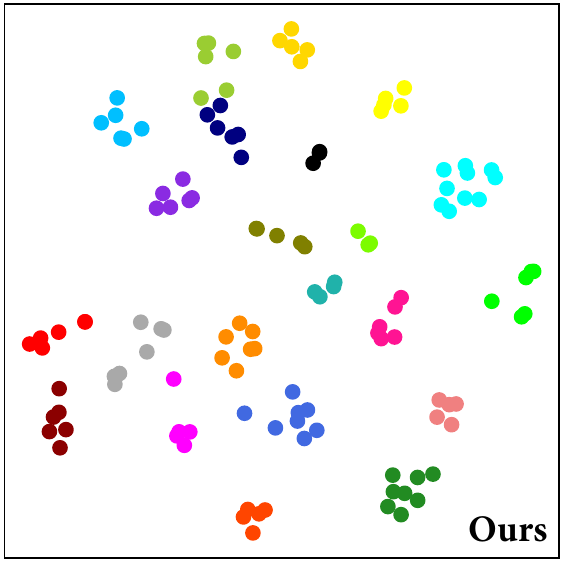} } 
 \subfigure[ Samples with Rotating Operation ]{ \includegraphics[width=0.24\linewidth]{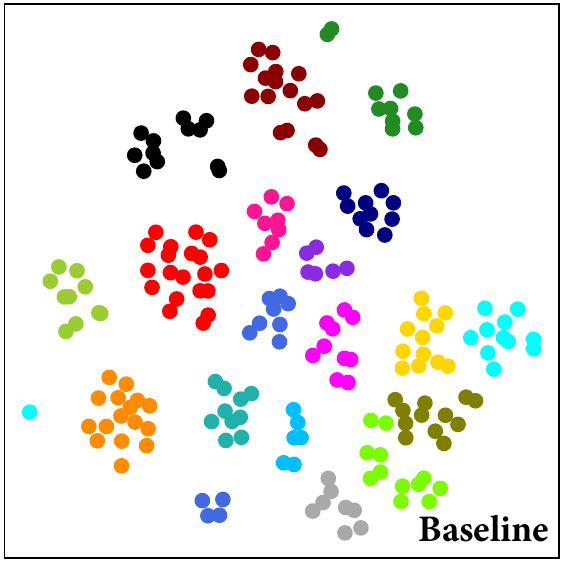} \includegraphics[width=0.24\linewidth]{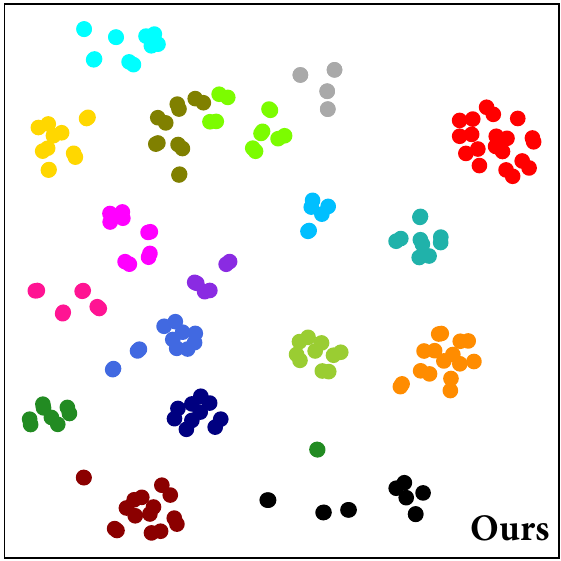}}
}
\caption{The feature distributions of our method and baseline under two different settings are compared.
\label{fig:visulization}}
  \vspace{-2mm}
\end{figure*}

\textbf{Effect of Rotation Invariant Constraint} The rotation invariant constraint strategy makes the network learning process more stable. The constraint relationship between the original feature and the rotated feature is established, which further avoids excessive bias caused by feature-level rotation. The rotation-invariant constraint is based on feature-level rotation, and experiments after adding this strategy show that mAP increased by 2.3\% and rank1 accuracy increased by 1.9\%. This shows that adding invariance constraints can effectively improve model performance.


\textbf{Parameter Experiments.} Fig.~\ref{fig:parameter} shows the two sets of parameter experiments we designed based on the feature level rotation. The numbers of rotated features denoted by $N$ are tested. The place where $N=0$ in the figure represents the result of the baseline. It can be found that the accuracy gradually increases as the number of rotated features increases. Since increasing the number of rotated features introduces more diversity, it will also increase the computational cost of the model. After $N$ reaches 5, the accuracy of the model no longer has a obvious upward trend. Besides, we evaluate the influence of the weights of the original feature represented by $\lambda$ and the rotated feature represented by $1-\lambda$. Experiments show that increasing the weight of the rotated feature or the weight of the original feature during the training process will make the model performance worse. The effect is best when the original feature and the rotated feature account for the same weight.

\begin{table}[t]
\centering
 \caption{\label{tab:test-ground}\small{TransReID \cite{he2021transreid} with our method embedded is evaluated on the two commonly used ReID datasets Market-1501 and MSMT17 based on city cameras. * represents the experimental results are reproduced by running their released code.}}
\begin{tabular}{p{2.8cm}p{1cm}p{1cm}p{1cm}p{1cm}}
\toprule
{Dataset} &\multicolumn{2}{c}{Market-1501}&\multicolumn{2}{c}{MSMT17} \\
\midrule
Method   & mAP  & $Rank1$ & mAP  & $Rank1$  \\
\midrule
\multicolumn{5}{l}{\textit{(a) Evaluation of CNN based methods}} \\
\midrule
MGN \cite{wang2018learning} &86.9 &  95.7 & 52.1 & 76.9 \\
OSNET \cite{zhou2019omni}  &84.9 & 94.8 & 52.9 & 78.7 \\
BagTricks \cite{luo2019strong}&85.9 &  94.5 & 45.1 & 63.4 \\
AGW \cite{ye2021deep} &87.8 &  95.1 & 49.3 & 68.3 \\
\midrule
\multicolumn{5}{l}{\textit{(b) Evaluation of Transformer based methods}}\\
\midrule
AAformer \cite{arXiv2021AAformer} & 87.7 & 95.4 &  63.2 & 83.6 \\
HAT \cite{MM2021HAT} & 89.5 & 95.6 &  61.2 & 82.3 \\
TransReID* \cite{he2021transreid}  & 89.5&  95.2  & 62.0 & 81.8 \\
\midrule
RotTrans (Ours)  & \bf{90.0} &	\bf{95.6} & \bf{67.4} & \bf{85.0} \\
\bottomrule
 \end{tabular}
   \vspace{-2mm}
\end{table}  

\textbf{Feature Distribution Comparison.} 
In order to show the rotation invariance more intuitively, we visualize the feature distribution under the two settings shown in Fig.~\ref{fig:visulization}. In detail, we use the widely-used t-SNE to reduce feature dimension to 2, and visualize dimension reduced feature in the 2D space.
\textit{a) Samples without Rotating Operation}:The first set of comparisons is the feature distribution of the model trained by our method and the model trained by the baseline on the same batch of test samples. The feature extracted by baseline has a relatively close distance between classes, while the feature extracted by our method is easy to distinguish. \textit{b) Samples with Rotating Operation}: The second group is about the further comparison of rotation invariance. On the basis of the first group, we perform random rotation on this batch of test samples when generating patch embeddings. In this setting, each test sample has both original and rotated forms. After introducing rotated samples, the performance of the baseline is degraded. However, it can be clearly seen that the rotated samples of the same class are closely clustered from the original samples by using our method.

\subsection{Evaluation on City Camera Datasets} 
The method we propose is not only suitable for UAV scenarios, but also performs well in general city camera scenarios. To demonstrate this, we conducted experiments as shown in Table \ref{tab:test-ground}. Benefiting from the advantage that our method can be arbitrarily embedded into transformer architectures, we can easily verify the performance of our method based on existing methods. specifically, we embed the proposed feature level rotation and invariance constraint strategies into TransReID \cite{he2021transreid}, and evaluate it on two commonly used person ReID datasets (Market-1501 \cite{zheng2015scalable} and MSMT17 \cite{wei2018person}) collected by city cameras. On the Market-1501 dataset, our method achieves 90\% mAP and 95.6\% rank1 accuracy, which outperforms most existing state-of-the-arts. Our method also performs well in the more complex data set MSMT17. The results show that our method also has strong generalization ability in common scenarios.

\section{Conclusion}
This paper proposes a transformer-based rotation invariance architecture, in order to solve the object recognition problem of large angle and direction changes in the data collected under the extreme viewing angle of the UAV. On multiple UAV datasets of pedestrians and vehicles, the evaluation results of the feature level rotation and invariance constraint strategies we designed for UAV scenario are much better than the existing state-of-the-arts. In addition, embedding our method into existing transformer based ReID models also achieves very competitive performance on city camera scene. We believe that our idea of rotation invariance at the feature level is not limited to ReID tasks, and it can be more generalized to many vision tasks with object rotation challenges in the future.

%
\begin{acks}
This work is supported by National Natural Science Foundation of China (62176188), Key Research and Development Program of Hubei Province (2021BAA187), Special Fund of Hubei Luojia Laboratory (220100015), the Bingtuan Science and Technology Program (No. 2019BC008) and CAAI-Huawei MindSpore Open Fund.
\end{acks}

\bibliographystyle{ACM-Reference-Format}
\bibliography{sample-base}










\end{document}